# Modelling and Path Planning of Snake Robot in cluttered environment

Akash Singh[1] , Chaohui Gong[2], Howie Choset[3]

*Abstract*—Studying snake robot locomotion in a cluttered environment has been a complicated and computationally expensive task. This is because the motion model is discontinuous due to the physical contact with obstacles, and the contact force cannot be determined solely by contact positions. We present a unique mathematical model of the robot interacting with obstacles in which the contact forces are mapped on the basis of a viscous friction model. Also a motion planning strategy has been introduced which helps deriving the simplest path that ensures sufficient number of contacts of the robot with the obstacles required to reach a goal position. Numerical simulations and experimental results are presented to validate the theoretical approach.

I. INTRODUCTION

Modular Snake robots[1,2,3] have shown their capability of moving through confined and narrow spaces, granular and muddy terrain, and pole systems. This versatility is achieved by a huge number of joints. Hence to study the locomotion and movement of such highly articulated robots, researchers have put a lot of effort in coordinating the motion of this hyper-redundant system. The previous works succeed in designing gait in a plain or organized environment, however, when it comes to unstructured environment obstacles, the gait fails in open environment. Researchers have shown that, in order to move in highly dense and complex environments snake robots tend to use the forces from obstacles when their body come in contact with the obstacles. Liljebäck et al in [4] focus on modelling and control of obstacle-aided motion of snake robot for locomotion in forward direction. Also their further work [5-7] introduces different mathematical models to formulate and model the obstacle aided motion of the snake robot. The research done in [4-7] is quite helpful in dynamically planning the motion of the snake robots but the complexity of the models is quite high. In this paper we present a solution to this complex problem that is less challenging to implement and more efficient computationally. This paper presents obstacle aided locomotion strategy for a modular snake robot [3] in a cluttered environment without entering into the dynamics of its motion.

The outline of this work is based upon two strategies: 1) Utilizing the forces from the obstacles for the forward propagation of robot without jumping into the dynamic details of its motion, since, the dynamic model becomes highly complex [8, 9] and also computationally expensive. 2) Making the snake robot follow the simplest path which is suitable for its complex body motion in the environment. This heuristic comes from the intuition that since of the snake robot locomotion is intricate, hence it would be easier for the snake robot to move through a path that it can follow without getting stuck between the obstacles or loosing its speed due to lack of contact with obstacles.

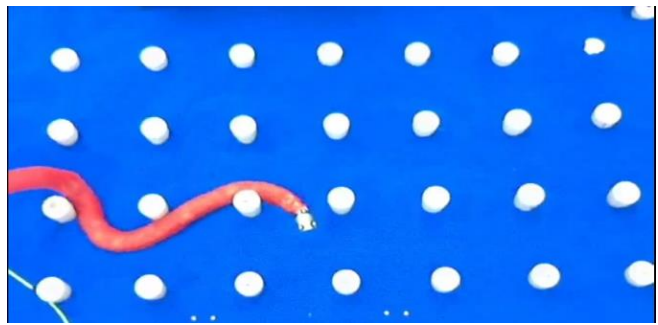

**Fig 1.** Modular Snake robot in viscous environment with cluttered Objects

In this paper firstly, we present a mathematical model for snake robot in a cluttered environment. The mathematical model is used to solve snake robot motion that relies on leveraging the non-dynamic effects of viscous forces on the locomotion snake robot in a swimming pool. This model predicts the behavior of the robot in a cluttered environment accurately which in turn helps in deriving the input angles to achieve a desired body velocity of the robot.

Another important aspect on which this research focusses is planning the path for the snake robot. Snake robots are often required to move to a desired target location in their on-field application in complex environments. Path planning of highly articulated robots is a tough challenge since the body frame is never stable. Another key challenge in planning the path for an obstacle aided motion is evaluating a test path. The evaluation should be based upon testing a path and choosing it such that it gives us a good set of contacts with rigid bodies. These contacts finally help the snake robot propel through the obstacles towards the goal position. In this paper, we provide a solution to both of the above mentioned problems. We use graphical voronoi diagrams as the planned paths to traverse through the dense obstacles and then use the depth search algorithm in the kinematic simulator to evaluate the test paths. The idea of using the GVG as the planned path relies on the fact that in densely cluttered environment the robot kinematics is not accurate and the robot tends to contact the pegs in order to move forward. Hence it utilizes these unavoidable contacts from the obstacle to move forward. For evaluating the GVG test paths we use the speed of the robot as the basic criteria. The speed serves as the simplest logical criteria since it takes into account the sufficient amount of contacts necessary to propel the robot. To validate our theory we have done simulations with the same mathematical model we have proposed and also applied the graphical path planning in the simulator. Experimental results have been included to show success of the model and planning method.

[1] Akash Singh is with Visvesvaraya National Institute of Technology, Nagpur (Akashvnit2016@gmail.com)
[2] Chaohui gong is with Carnegie Mellon University and Inbitorobotics (chaohuigong@bitorobotics.com)
[3] Professor at Robotics Institute, Carnegie Mellon University (choset@cs.cmu.edu)

## II. SNAKE ROBOT MODEL

The derivation of the N-link snake model is based upon the viscous friction model as described in [10]. The reason for this approach is to imitate the motion of a biological snake in viscous environment. Before modelling the snake robot we take two assumptions 1) The environment is highly dense with obstacles so there are enough contacts that minimize slip of the snake robot's locomotion. 2) Biological snake motions are dominated by friction of the viscous environment and any inertial/dynamic effect is immediately dissipated. The second characteristic is reminiscent of the two fundamental assumptions for systems swimming in low Reynolds number fluid [11]. They are given by: 1) The resistive force of a rigid body is proportional to its body velocity (viscosity assumption) and 2) The total net forces experienced by a system sum to zero (dynamic dissipation). Therefore the snake robot is modelled in a way similar to a low-Reynolds number swimmer [8].

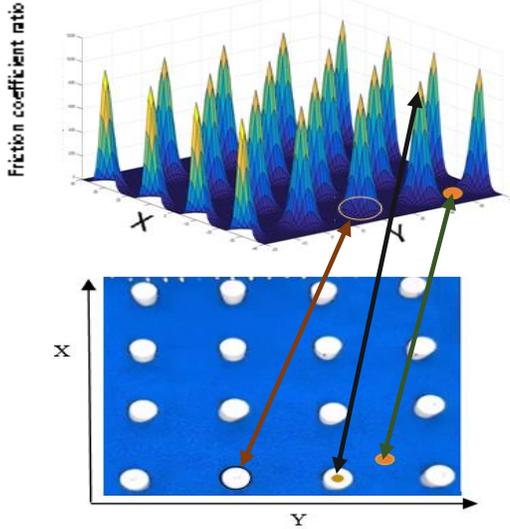

**Fig 1.2**: Shows the equivalence of peaks of the Gaussian hump with the center of the obstacles by blue arrow; the low level surface of Gaussian map to the granular surface by green arrow; the thickness of the pegs and the thickness Gaussian hump by brown arrow.

The viscous friction model is similar to [5] instead of scheduling contacts we consider the whole body of the snake robot to be planar and in contact with ground. The body velocity of link i with respect to the head frame (link 0) is denoted as $\xi_{0i}$ and is computed using the differential mapping,

$$\xi_{0i} = J_i^b [\ \dot{\alpha}_1\ \dot{\alpha}_2\ \cdots\ \dot{\alpha}_i\ ]', \quad (1)$$

$J_i^b \in R^{3 \times i}$ is the body Jacobian when viewing link i as the tool frame[18] and α denotes the internal joint angles(shape variables). To facilitate further deviation we rewrote (1) into the following,

$$\xi_{0i} = [J_i^b\ \ 0_i]\ \dot{\alpha} = J_i \dot{\alpha}, \quad (2)$$

where $0_i \in R^{3 \times (N-i)}$ denotes a matrix filled with zeros and $J_i \in R^{3 \times N}$ is a linear differential map from joint velocity (shape velocity) $\dot{\alpha}$ to $\xi_{0i}$. We denote the body velocity of the head with respect to the world frame as $\xi_{w0}$. The body velocity of individual links with respect to the world frame is then computed using the following [6],

$$\xi_{wi} = Ad_{g_{oi}^{-1}} \xi_{w0} + \xi_{0i}, \quad (3)$$

Where $Ad_g$ denotes the adjoint operator, which maps body velocity between different frames. The viscous force acting on a link in contact with an obstacle is larger in the lateral direction as compared to the longitudinal direction. The ratio of viscous force due to every obstacle at a link in the lateral direction to the longitudinal direction is modeled as a 2D Gaussian hump. The viscous force experienced by link i with respect to its own body frame was thus computed as

$$F_i^b = -K_i \xi_{wi} = -\begin{bmatrix} k_{xi} & & \\ & k_{yi} & \\ & & k_{zi} \end{bmatrix} \xi_{wi}, \quad (4)$$

Where $k_{xi} = k_{zi} = 1$, $k_{yi}$ = Viscous friction value attained at the center of mass of the $i^{th}$ link by calculating the sum of all the viscous friction coefficient due to all the obstacles in the environment. The value of $k_{yi}$ for a link with its COM position being X, Y is computed by the following:

$$K_{yi} = K\_height + \sum A * \exp\left(-\left(\frac{(x-x_k)^2}{2\sigma_x^2} + \frac{(y-y_k)^2}{2\sigma_y^2}\right)\right) \quad (5),$$

N is the number of cluttered obstacles present in the environment. $x_k, y_k$ represent the position of center of the $k_{th}$ obstacle. The above expression states that if any of the snake's link is near to any obstacle the coefficient of viscous friction increases massively in the lateral direction. Fig 1.2(top) shows the viscous friction ratio in the lateral direction to the longitudinal direction at the snake links accurately in a uniformly placed obstacle environment. The arrows in Fig 1.2 show the corresponding relationship between the obstacle (in white) cluttered environment and the equivalent Viscous friction coefficient ratio map. The peaks represent the higher ratio due to obstacles and the flat plane represent the small viscous forces due to ground in the lateral direction to the longitudinal direction. The $F_i^b$ is then transformed in the head frame using the following [7],

$$F_i = Ad'_{g_{oi}} F_i^b, \quad (6)$$

Where $F_i$ denotes the force applied to link i with respect to the head frame. The low Reynolds number assumption implies that all the dynamic effects are dissipated during locomotion. Therefore, the net forces experienced by the system sums to zero [8],

$$\sum_{i=0}^{N} F_i = 0, \qquad (7)$$

Rewriting (7) in terms of $\dot{\alpha}$ and $\xi_{w0}$ led to

$$\sum_{i=0}^{N} \left( Ad_{g_{oi}} K_i \left( Ad_{g_{oi}}^{-1} \xi_{w0} + J_i \dot{\alpha} \right) \right) = 0, \qquad (8)$$

The only unknown variable i.e the body velocity of the head frame, $\xi_{w0}$. Further manipulation, resulted in the following,

$$\left( \sum_{i=0}^{N} \left( Ad_{g_{oi}}' K_i Ad_{g_{oi}}^{-1} \right) \right) \xi_{w0} = - \sum_{i=0}^{N} \left( Ad_{g_{oi}}' K_i J_i \right) \dot{\alpha} \qquad (9)$$

The body velocity of the head frame is computed as

$$\xi_{w0} = -\omega_\xi^{-1} \omega_\alpha \dot{\alpha} = A(\alpha)\dot{\alpha}, \qquad (10)$$

which was determined by instantaneous shape changes. The above relation was used to simulate the snake robot in the simulator. The simulation was done in time steps of time interval 0.05 seconds.

### III. PLANNING

The idea of introduction to planning in to the simulation is to find the easiest path for the snake robot to reach a particular target point in the environment. The environment consists of the N link snake robot and randomly placed identical round obstacles placed in the environment. This helps planning a path the snake robot can easily follow. By "easy to follow" term we mean that we choose a path in which the snake robot doesn't get stuck in the obstacles or the magnitude of velocity of center of mass doesn't fall below a certain limit at any particular point while following the path.

*Generalised Voronoi Graph Path Plan*

GVG path planner was chosen so as to minimize the complexity of choosing the exact path and also ensure the path chosen justifies the motion requirements of the snake robot. The idea behind using GVG is following: 1) Every GVG branch is such that it ensures a segment of the path that avoids possibility of collision with any obstacle as much as possible while traversing that segment. 2) Every branch of GVG represents a set of homotopic paths. The first idea is valid and ensures success only if the environment densely cluttered enough. Hence idea relies on not necessarily for the snake robot to follow the exact voronoi path rather it can follow any of the path in the homotopic class represented by voronoi path. This is achieved by commanding the robot head to point towards the target point on the path. Since the robot does not follow the voronoi path accurately rather it follows a deformed curve of the voronoi branch. This deformed curve is homotopic to the straight branch which is the segment of the trial path. As stated earlier the environment consists of highly dense obstacles. This is stated such that each voronoi branch becomes the heuristic for good contacts for the snake robot with the obstacles. By good contacts we mean the set of contacts of the snake robot with the obstacles that help it propel forward and a super set of such sets gives the complete set of contacts that help the robot traverse completely through the cluttered path. In the previous works such as [4, 5, 6] the path chosen by the snake robot is completely dependent on the dynamics of the robot acted upon by the forces and the series of contact that lead the snake robot to traverse through the cluttered environment. But our idea here is to make the snake robot reach to a particular target point using the easiest path and avoid the path that difficult to follow or that are unfavourable for the snake robot to reach without losing its speed in between. The target point as mentioned above is a point belonging to the set of point in the voronoi graph set.

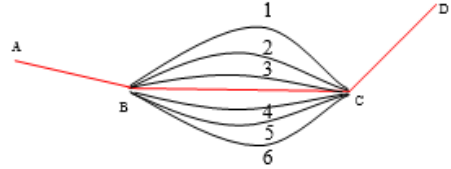

**Fig 1.3** shows the set of paths (1-6) homotopic to a-c

The above diagram shows some of the paths 1, 2, 3, 4, 5, 6 that are homotopic to the segment BC. Here AB, BC and CD are the branches of the voronoi graph and segments of the current trial path.

### IV. PATH FOLLOWING STRATAEGY OF SNAKE ROBOT

A little research [9] has been done on path following of limbless snake robots. The paper introduces the path following strategy for the planar snake robot using a line of sight guidance law. The animated snake robot as described below in Fig1.4 is made to follow the voronoi branch using a simple strategy. The head of the snake robot is directed to point towards a temporary target point on the present path. The temporary target point is chosen on the basis of line of sight mechanism in which the robot's head module is pointed. This line of sight target point is achieved by getting the point of intersection of a circle centered at the robot head module and the voronoi branch line. This angle value then travels through the whole snake robot body, that is the all individual robot modules attain that angle at wave kind of pattern. Similarly, the subsequent angle attained by head travels through all the links after certain time interval. This is the concept of travelling wave. The wave travels throughout the body of the snake robot. The snake robot's body hence moves towards the target point if there is minimal slipping, which was assumed at the very beginning of this paper. The angle attained by the head is given by a proportional and differential controller as described by eq(11). The input angle ($\Delta\theta$) is calculated by the angle between the line joining the center head module to the target point and the line represented by the head vector direction.

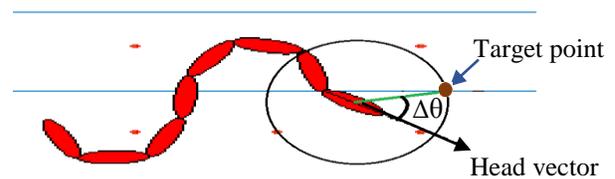

**Fig 1.4** shows snake robot's head vector at an angle of $\Delta\theta$ with the straight line path from the snake head to the temporary target point.

The closed loop head angle control is given by:
Input angle to the head motor = $P*(\Delta\theta) + K*(\Delta\theta')$;     (11)

## V. Graph Search Algorithm

The graph search algorithm helps finding the ideal path for the snake robot in order to traverse through the cluttered environment. Generalized voronoi graph of the test environment consists of the straight branches existing at equal distances from the obstacles joining the vertices. A set of such consequent vertices of fixed length make up a particular shape or configuration and is known as a node. In the simulator the snake robot is commanded to follow the current branch that is joining the adjacent vertices and subsequently enhancing the graph by adding up the next vertex to the node. The robot progresses towards the goal following a certain set of nodes and when it crosses a target vertex, it chooses the next target vertex by depth search and it adds up the next vertex to the currently formed path. The extension of the formed path depends upon the ease with which the snake robot travels from the previous vertex to the next. Suppose a situation arises that the robot speed reduces below a certain threshold, it backtracks to the previous vertex and takes up a different vertex as its target vertex in the second attempt according to the depth search algorithm. While doing so it reduces the previously formed path by discarding the last node and testing a new node consisting of the new target vertex. This process is iterated until the robot reaches the target. Finally the planned path is formed by connecting all the target vertices set through which the robot passed successfully.

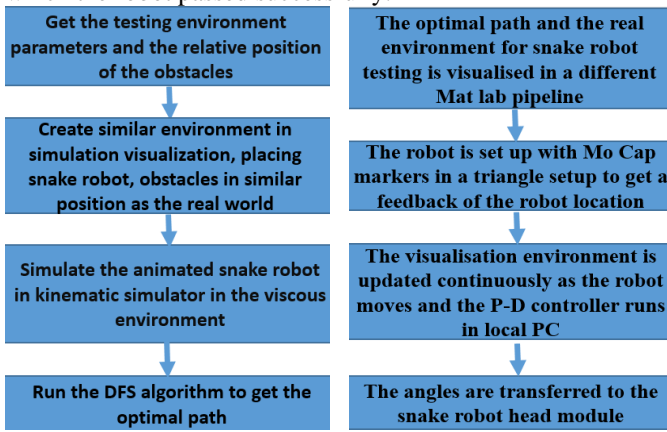

**Fig 1.5**: *1) Flow Chart in left shows the steps of getting optimal path via simulation. 2) Flow chart in right shows sequence for testing the path in real time snake*

## VI. Simualations and Experimental results

Fig 1.5 1) shows the flowchart of retrieving the planned path using depth search algorithm and 2) shows the experimental flow chart to test the real snake robot in the testing environment. The simulation was done for the 9 link snake robot in a viscous material by keeping the viscous ratio ky/kx as 4 as chosen on the basis of previous experimentations and observations of the snake robot in the granular pool, ratio of the peg spacing to the wavelength being 4. The obstacles used here are represented by identical round pegs of 4 inch diameter. To create the similar kind of environment in the simulations the values of the sigma_x and sigma_y was adjusted so that the hump of the 2d gaussian attains a width of 4 inches, which is the actual width of the pegs used in the real environment. One of the important observations while simulating was the variation of average speed of the snake robot with the density of the obstacles. It is noted that the speed of the snake robot increases with increase in the density of the pegs representing the obstacles. In the figure 1.5 given below the red dots represent the center of the obstacles and the blue graph is the generated voronoi graph for the available obstacle placement configuration.

The simulations were performed for the above mentioned mathematical model on MATLAB and ideal paths were retrieved for the 9 link snake robot for different peg configurations. Various paths were generated for different peg placements in the simulator. The path generated after simulations were tested on the real snake robot and with the same peg placements in the real environment. For the experiments we use the closed loop control of the snake robot using a set of four Motion capture cameras. To get the position of the snake robot we use a set of three mocap markers placed on the head (the first link) of the snake robot. The camera's software returns the 3 dimensional positions of the three markers, the markers are placed in a triangular position with each other. By getting the position of the three markers we calculate the position as well as the orientation of the head of the snake robot. The head of the snake robot is turned towards the target point on the path drawn on the visualization map. The visualization map code was written in matlab that links the Mocap markers input data with the position and orientation of the snake robot. The reference angle given to the snakes head is calculated using the similar calculation as we did to control the head of the snake in the simulator. The target point is choosen with the help of the visualization map in MATLAB as shown in the figure. The desired angle given to the snakes head is using a closed loop PD controller and the angle travels along the body of the snake. Figures 1.6 and 1.7 show experimental and simulation results of the snake robot tending to follow the path shown by black arrow in figures (a-b) and (e-f).The set of figures clearly show the slightly

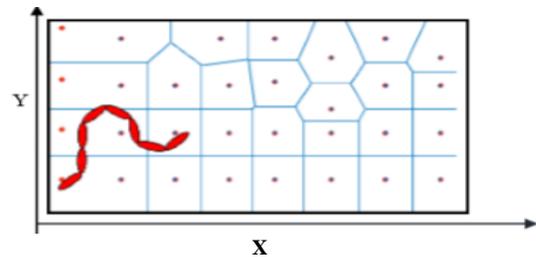

**Fig 1.6**: Kinematic Simulation environment dots representing obstacles centers, blue lines representing Voronoi graph.

different paths planned for two slightly differently cluttered environments. The only significant difference being the distance d between the red and blue marked pegs. The simulation was carried out for the two environments as shown:

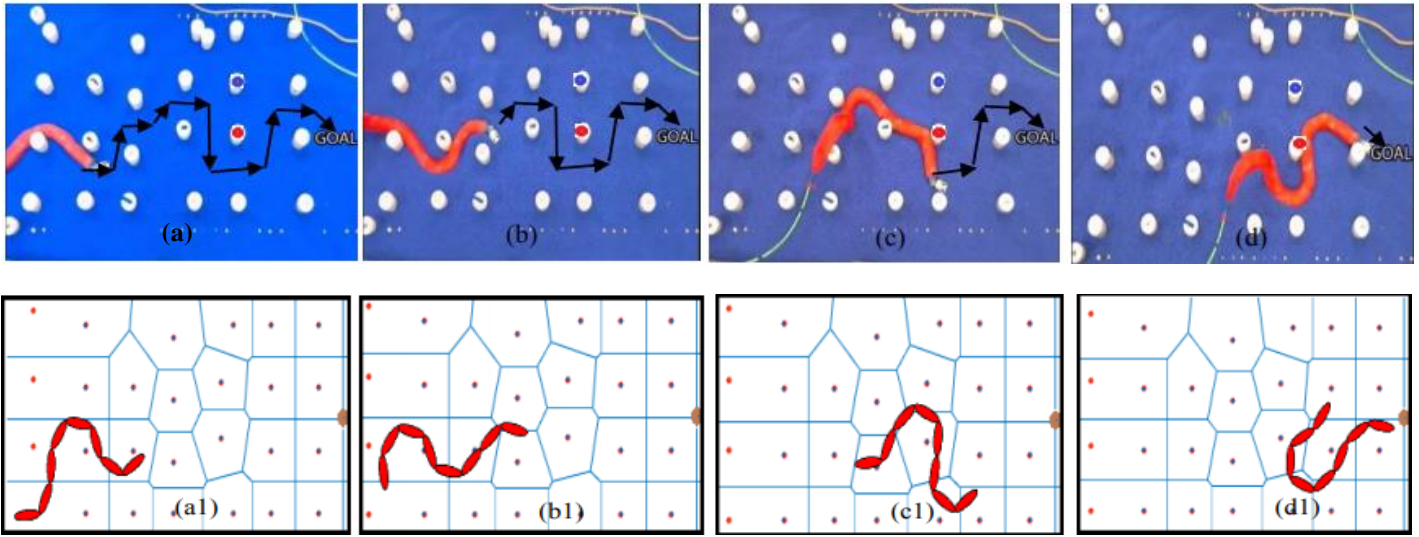

**Fig 1.6** Stages (a-d) show the robot travelling through the environment for the distance b/w colored pegs d=10inches.
Stages (a1-d1) show the corresponding simulation results, while displaying the voronio graph of the environment (brown patch is the goal).

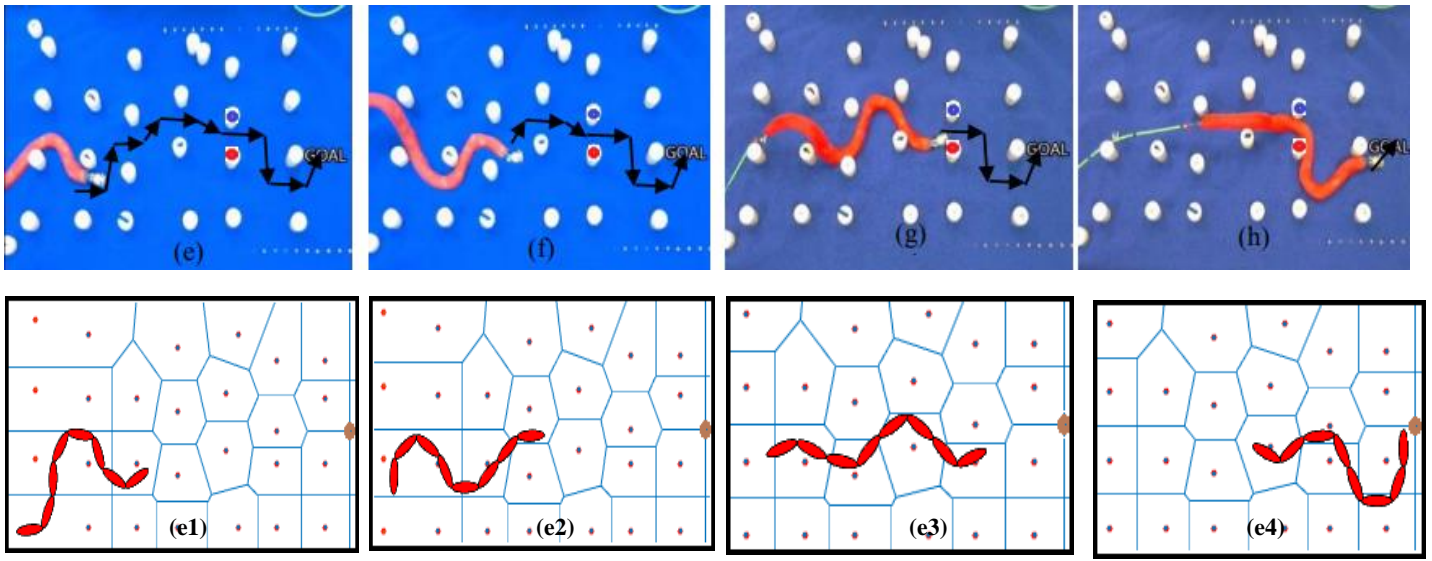

**Fig 1.7** Stages (e-f) show the robot travelling through the environment for the distance b/w colored pegs d=8 inches.
Stages (e1-f1) show the corresponding simulation results, while displaying the voronio graph of the environment (brown patch is the goal)

## VII. REPEATABILITY

The snake robot was set to follow a simulated path in real environment. One of the important factors for the success of any experiment is the repeatability to perform a certain task. Experimentations were done on the snake robot making it follow a particular simulated and planned path in the real environment. The experiment was repeated 25 times. Each trial was regarded successful if the snake robot followed that path to reach the goal position. Out of 25 trials the robot was successful 20 times. Hence the repeatability of the robot was calculated to be 0.8. Along with repeatability other experiments were done on new paths. Along with a good repeatability, the robot also has a high success rate in following new paths in the first trial keeping the controller parameters fixed. But in some uniquely simulated paths the robot sometimes fails to follow the path at the first attempt. Hence, some of the parameters need to be tuned in order to make the path following more efficient. The parameters that need to be tuned for such tasks include the frequency of the travelling wave, P and D constants for the head controller. Though the tuning doesn't vary much around the ideal value of these parameters, hence the tuning doesn't consume much time. Once these parameters are set again, the robot was set to follow the new paths, and yet again the robot had high repeatability in following those paths.

## VIII. CONCLUSION AND FUTURE WORKS

This paper presented a mathematical model for the snake robot interacting with obstacles in a cluttered environment where the obstacle contact forces were modelled to anisotropic viscous friction. The model was tested in the simulations successfully. The model helps analysing the robot motion in such complex environment using system kinematics. Graphical Voronoi graph was used as a tool to plan the optimal path of the environment, consisting of obstacles as the sites of the graph and the bisecting lines as test paths. The test paths were evaluated in the simulator considering speed of the robot as the primary criteria. The experimental results prove the correctness of the approach. A repeatability of 0.8 of following a planned path by the snake robot shows the idea of choosing the GVG graph as the test path and assuming the branches of the graph as the representative of all the homotopic paths correct. The high repeatability value proves the correctness of the method of developing a simplified kinematic model of the robot in such complex environments. This ensures the low computation cost of planning the optimal path for the robot in complex and cluttered environment.

Future work will focus upon testing the derived mathematical model and planning strategy in more complicated environments and different viscous environments. Apart from this the experimental parameter values will be generalized if possible, the other planning algorithms like Depth first search and Dijkstra algorithms for getting optimal paths could be tested in such cluttered environments.